\documentclass[12pt]{article}
\usepackage{amsmath}
\usepackage{amssymb}
\usepackage{amsfonts}

\input{tcilatex}
\begin{document}

\begin{center}
{\LARGE Linear Classification of data with Support Vector Machines and
Generalized Support Vector Machines}{\large \medskip }

Xiaomin\ Qi, Sergei Silvestrov and Talat Nazir\bigskip

{\small \textit{Division of Applied Mathematics, School of Education,}}

{\small \textit{Culture and Communication, M\"{a}lardalen University, 72123 V%
\"{a}ster\aa s, Sweden.}}

E-mail{\small : }xiaomin.qi@mdh.se, sergei.silvestrov@mdh.se,
talat.nazir@mdh.se\bigskip\ 
\end{center}

\noindent
--------------------------------------------------------------------------------------------

\noindent \textit{Abstract:} \ \ In this paper, we study the support vector
machine and introduced the notion of generalized support vector machine for
classification of data. We show that the problem of generalized support
vector machine is equivalent to the problem of generalized variational
inequality and establish various results for the existence of solutions.
Moreover, we provide various examples to support our results.

\noindent \textbf{---------------------------------------------}



\noindent \textbf{---------------------------------------------}

\section{Support Vector Machine}

Over the last decade, support vector machines (SVMs) \cite{CV95, CS2000,
V96, V98, WW98} has been revealed as very powerful and important tools for
pattern classification and regression. It has been used in various
applications such as text classification \cite{J98}, facial expression
recognition \cite{MK2003}, gene analysis \cite{GWB2002} and many others \cite%
{AC09, KKAB12, LKZW03, LWBM08, N2004, SCD14, SL12, WLY08, WWB11, XMSW10,
ZLC14, ZC11}. Recently, Wang et al. \cite{WQWD} presented SVM based fault
classifier design for a water level control system. They also studied the
SVM classifier based fault diagnosis for a water level process \cite{WQWDO14}%
.

For the standard support vector classification (SVC), the basic idea is to
find the optimal separating hyperplane between the positive and negative
examples. The optimal hyperplane may be obtained by maximizing the margin
between two parallel hyperplanes, which involves the minimization of a
quadratic programming problem.

Support Vector Machines is based on the concept of decision planes that
define decision boundaries. A decision plane is one that separates between a
set of objects having different class memberships.

Support Vector Machines can be thought of as a method for constructing a
special kind of rule, called a linear classifier, in a way that produces
classifiers with theoretical guarantees of good predictive performance (the
quality of classification on unseen data).\medskip

In this paper, we study the problems of support vector machine and define
generalized support vector machine. We also show the sufficient conditions
for the existence of solutions for problems of generalized support vector
machine. We also support our results with various examples.\medskip

Thought this paper, by $%
\mathbb{N}
,$ $%
\mathbb{R}
,$ $%
\mathbb{R}
^{n}$ and $\mathbb{R}_{n}^{+}$ we denote the set of all natural numbers, the
set of all real numbers, the set of all $n$-tuples real numbers, the set of
all $n$-tuples of nonnegative real numbers, respectively.

Also, we consider $\left\Vert \cdot \right\Vert \ $and $<\cdot ,\cdot >$ as
Euclidean norm and usual inner product on $%
\mathbb{R}
^{n}$, respectively.

Furthermore, for two vectors $\mathbf{x,y\in }$ $%
\mathbb{R}
^{n},$ we say that $\mathbf{x\leq y}$ if and only if $x_{i}\leq y_{i}$ for
all $i\in \{1,2,...,n\},$ where $x_{i}$ and $y_{i}$ are the components of $%
\mathbf{x}$ and $\mathbf{y},$ respectively.\medskip

\noindent {\Large Linear Classifiers}\newline

\noindent Binary classification is frequently performed by using a function $%
f:%
\mathbb{R}
^{n}\rightarrow 
\mathbb{R}
$ in the following way: the input $\mathbf{x}=\left( x_{1},...,x_{n}\right) $
is assigned to the positive class if, $f\left( \mathbf{x}\right) \geq 0$ and
otherwise to the negative class. We consider the case where $f\left( \mathbf{%
x}\right) $ is a linear function of $\mathbf{x},$ so that it can be written
as%
\begin{eqnarray*}
f\left( \mathbf{x}\right) &=&\left\langle \mathbf{w}\cdot \mathbf{x}%
\right\rangle +b \\
&=&\sum_{i=1}^{n}w_{i}x_{i}+b,
\end{eqnarray*}%
where $\mathbf{w}\in 
\mathbb{R}
^{n},b\in 
\mathbb{R}
$ are the parameters that control the function and the decision rule is
given by $sgn\left( f\left( \mathbf{x}\right) \right) .$ The learning
methodology implies that these parameters must be learned from the
data.\bigskip

\noindent \textbf{Definition 1.1. }\ \ We define the functional margin of an
example $\left( \mathbf{x}_{i},y_{i}\right) $ with respect to a hyperplane $%
\left( \mathbf{w},b\right) $ to be the quantity 
\begin{equation*}
\gamma _{i}=y_{i}\left( \left\langle \mathbf{w}\cdot \mathbf{x}%
_{i}\right\rangle +b\right) ,
\end{equation*}%
where $y_{i}\in \{-1,1\}.$ Note that $\gamma _{i}>0$ implies correct
classification of $\left( \mathbf{x}_{i},y_{i}\right) .$ If we replace
functional margin by geometric margin we obtain the equivalent quantity for
the normalized linear function $\left( \frac{1}{\left\Vert \mathbf{w}%
\right\Vert }\mathbf{w},\frac{1}{\left\Vert \mathbf{w}\right\Vert }b\right)
, $ which therefore measures the Euclidean distances of the points from the
decision boundary in the input space.

\noindent Actually geometric margin can be written as 
\begin{equation*}
\tilde{\gamma}=\frac{1}{\left\Vert \mathbf{w}\right\Vert }\gamma .
\end{equation*}%
To find the hyperplane which has maximal geometric margin for a training set 
$S$ means to find maximal $\tilde{\gamma}.$ For convenience, we let $\gamma
=1,$ the objective function can be written as 
\begin{equation*}
\max \frac{1}{\left\Vert \mathbf{w}\right\Vert }.
\end{equation*}%
Of course, there have some constraints for the optimization problem.
According to the definition of margin, we have $y_{i}\left( \left\langle 
\mathbf{w}\cdot \mathbf{x}_{i}\right\rangle +b\right) \geq 1,$ $i=1,...,l.$
We rewrite the equivalent formation of the objective function with the
constraints as 
\begin{equation*}
\min \frac{1}{2}\left\Vert \mathbf{w}\right\Vert ^{2}\quad \text{such that }%
\quad y_{i}\left( \left\langle \mathbf{w}\cdot \mathbf{x}_{i}\right\rangle
+b\right) \geq 1,i=1,...,l.
\end{equation*}%
We denote this problem by SVM.

\section{Generalized Support Vector Machines}

\noindent We replace $\mathbf{w},b$ by $W,B$ respectively, the control
function $F:%
\mathbb{R}
^{n}\rightarrow 
\mathbb{R}
^{n}$ defined as 
\begin{equation}
F\left( \mathbf{x}\right) =W.\mathbf{x}+B,  \tag{2.1}
\end{equation}%
where $W\in 
\mathbb{R}
^{n\times n},$ $B\in 
\mathbb{R}
^{n}$ are the parameters of control function.

\noindent Define%
\begin{equation}
\tilde{\gamma}_{k}^{\ast }=\mathbf{y}_{k}\left( W\mathbf{x}_{k}+B\right)
>1\quad \text{for}\quad k=1,2,...,l,  \tag{2.2}
\end{equation}%
where $\mathbf{y}_{k}\in \left\{ \left( -1,-1,...,-1\right) ,\left(
1,1,...,1\right) \right\} $ is $n$ dimensional vector.\bigskip

\noindent \textbf{Definition 2.1.} \ \ We define a map $G:%
\mathbb{R}
^{n}\rightarrow 
\mathbb{R}
_{+}^{n}$ by%
\begin{equation}
G\left( \mathbf{w}_{i}\right) =\left( \left\Vert \mathbf{w}_{i}\right\Vert
,\left\Vert \mathbf{w}_{i}\right\Vert ,...,\left\Vert \mathbf{w}%
_{i}\right\Vert \right) \quad \text{for}\quad i=1,2,...,n,  \tag{2.3}
\end{equation}%
where $\mathbf{w}_{i}$ be the row of $W_{n\times n}$ for $i=1,2,...,n.$

The problem is find $\mathbf{w}_{i}\in 
\mathbb{R}
^{n}$ that satisfy%
\begin{equation}
\min_{\mathbf{w}_{i}\in W}G\left( \mathbf{w}_{i}\right) \text{ such that }%
\eta >0,  \tag{2.4}
\end{equation}%
where $\eta =\mathbf{y}_{k}\left( W\mathbf{x}_{k}+B\right) -1.$\newline

\noindent We call this problem as the Generalized Support Vector Machine
(GSVM).

The GSVM is equivalent to%
\begin{equation*}
\text{find }\mathbf{w}_{i}\in W:\text{\ }\left\langle G^{\prime }\left( 
\mathbf{w}_{i}\right) ,\mathbf{v}-\mathbf{w}_{i}\right\rangle \geq 0\quad 
\text{for all }\mathbf{v}\in 
\mathbb{R}
^{n}\text{ with }\eta >0,
\end{equation*}%
or more specifically%
\begin{equation}
\text{find }\mathbf{w}_{i}\in W:\text{\ }\left\langle \eta G^{\prime }\left( 
\mathbf{w}_{i}\right) ,\mathbf{v}-\mathbf{w}_{i}\right\rangle \geq 0\quad 
\text{for all}\ \mathbf{v}\in 
\mathbb{R}
^{n}.  \tag{2.5}
\end{equation}%
Hence the problem of GSVM becomes to the problem of generalized variational
inequality.\bigskip

\noindent \textbf{Example 2.2.}\ \ \ Let us take the group of points
positive class $\left( 1,0\right) ,$ $\left( 0,1\right) $ and negative class 
$\left( -1,0\right) $, $\left( 0,-1\right) .$

\noindent First we use \underline{SVM} to solve this problem to find the
hyperplane $<\mathbf{w},\mathbf{x}>+b=0$ that separate this two kinds of
points. Obviously, we know that the hyperplane is $H$ which is shown in the
Figure.%
\begin{equation*}
\FRAME{itbpF}{3.243in}{2.6991in}{0in}{}{}{Figure}{\special{language
"Scientific Word";type "GRAPHIC";maintain-aspect-ratio TRUE;display
"USEDEF";valid_file "T";width 3.243in;height 2.6991in;depth
0in;original-width 3.1981in;original-height 2.6558in;cropleft "0";croptop
"1";cropright "1";cropbottom "0";tempfilename
'NQCS3C00.wmf';tempfile-properties "XPR";}}
\end{equation*}

For two positive points, we have%
\begin{eqnarray*}
\left( w_{1},w_{2}\right) \left[ 
\begin{array}{c}
1 \\ 
0%
\end{array}%
\right] +b &=&1 \\
\left( w_{1},w_{2}\right) \left[ 
\begin{array}{c}
0 \\ 
1%
\end{array}%
\right] +b &=&1
\end{eqnarray*}%
which implies%
\begin{eqnarray*}
w_{1}+b &=&1 \\
w_{2}+b &=&1.
\end{eqnarray*}

\noindent For two negative points, we have%
\begin{eqnarray*}
\left( w_{1},w_{2}\right) \left[ 
\begin{array}{c}
-1 \\ 
0%
\end{array}%
\right] +b &=&-1 \\
\left( w_{1},w_{2}\right) \left[ 
\begin{array}{c}
0 \\ 
-1%
\end{array}%
\right] +b &=&-1
\end{eqnarray*}%
implies that%
\begin{eqnarray*}
-w_{1}+b &=&-1 \\
-w_{2}+b &=&-1.
\end{eqnarray*}

\noindent From the equations, we get $\mathbf{w}=\left( 1,1\right) $ and $%
b=0 $. The result is $\left\Vert \mathbf{w}\right\Vert =\sqrt{2}$.

\bigskip

\noindent Now we apply \underline{GSVM} for this data.

\noindent For two positive points, we have%
\begin{equation*}
\left[ 
\begin{array}{cc}
w_{11} & w_{12} \\ 
w_{21} & w_{22}%
\end{array}%
\right] \left[ 
\begin{array}{c}
1 \\ 
0%
\end{array}%
\right] +\left[ 
\begin{array}{c}
b_{1} \\ 
b_{2}%
\end{array}%
\right] =\left[ 
\begin{array}{c}
1 \\ 
1%
\end{array}%
\right]
\end{equation*}%
and%
\begin{equation*}
\left[ 
\begin{array}{cc}
w_{11} & w_{12} \\ 
w_{21} & w_{22}%
\end{array}%
\right] \left[ 
\begin{array}{c}
0 \\ 
1%
\end{array}%
\right] +\left[ 
\begin{array}{c}
b_{1} \\ 
b_{2}%
\end{array}%
\right] =\left[ 
\begin{array}{c}
1 \\ 
1%
\end{array}%
\right]
\end{equation*}%
which gives%
\begin{equation}
\left[ 
\begin{array}{c}
w_{11} \\ 
w_{21}%
\end{array}%
\right] +\left[ 
\begin{array}{c}
b_{1} \\ 
b_{2}%
\end{array}%
\right] =\left[ 
\begin{array}{c}
1 \\ 
1%
\end{array}%
\right] \ \text{and }\left[ 
\begin{array}{c}
w_{12} \\ 
w_{22}%
\end{array}%
\right] +\left[ 
\begin{array}{c}
b_{1} \\ 
b_{2}%
\end{array}%
\right] =\left[ 
\begin{array}{c}
1 \\ 
1%
\end{array}%
\right] .  \tag{2.6}
\end{equation}

For two negative points, we have%
\begin{equation*}
\left[ 
\begin{array}{cc}
w_{11} & w_{12} \\ 
w_{21} & w_{22}%
\end{array}%
\right] \left[ 
\begin{array}{c}
-1 \\ 
0%
\end{array}%
\right] +\left[ 
\begin{array}{c}
b_{1} \\ 
b_{2}%
\end{array}%
\right] =\left[ 
\begin{array}{c}
-1 \\ 
-1%
\end{array}%
\right]
\end{equation*}%
and%
\begin{equation*}
\left[ 
\begin{array}{cc}
w_{11} & w_{12} \\ 
w_{21} & w_{22}%
\end{array}%
\right] \left[ 
\begin{array}{c}
0 \\ 
-1%
\end{array}%
\right] +\left[ 
\begin{array}{c}
b_{1} \\ 
b_{2}%
\end{array}%
\right] =\left[ 
\begin{array}{c}
-1 \\ 
-1%
\end{array}%
\right] ,
\end{equation*}%
which provides%
\begin{equation}
\left[ 
\begin{array}{c}
-w_{11} \\ 
-w_{21}%
\end{array}%
\right] +\left[ 
\begin{array}{c}
b_{1} \\ 
b_{2}%
\end{array}%
\right] =\left[ 
\begin{array}{c}
-1 \\ 
-1%
\end{array}%
\right] \text{ and }\left[ 
\begin{array}{c}
-w_{12} \\ 
-w_{22}%
\end{array}%
\right] +\left[ 
\begin{array}{c}
b_{1} \\ 
b_{2}%
\end{array}%
\right] =\left[ 
\begin{array}{c}
-1 \\ 
-1%
\end{array}%
\right] .  \tag{2.7}
\end{equation}%
From (2.6) and (2.7), we get%
\begin{equation*}
W=\left[ 
\begin{array}{cc}
1 & 1 \\ 
1 & 1%
\end{array}%
\right] \quad B=\left[ 
\begin{array}{c}
b_{1} \\ 
b_{2}%
\end{array}%
\right] =\left[ 
\begin{array}{c}
0 \\ 
0%
\end{array}%
\right] .
\end{equation*}%
Thus%
\begin{equation*}
\min G\left( \mathbf{w}_{i}\right) =\min \left\{ G\left( \mathbf{w}%
_{1}\right) ,G\left( \mathbf{w}_{2}\right) \right\} =(\sqrt{2},\sqrt{2}).
\end{equation*}%
Hence we get $\mathbf{w}=\left( 1,1\right) $ that minimize $G\left( \mathbf{w%
}_{i}\right) $ for $i=1,2.$ $\square $\bigskip

\noindent \textbf{Conclusion:} \textbf{\ \ }The above example shows that we
get same result by applying any method SVM and GSVM.\bigskip

In the next example, we consider the two distinct group of data, first solve
both data for separate cases and then solve it for combine case for both
methods SVM and GSVM.

\noindent \textbf{Example 2.3.} \ \ Let us consider the three categories of
data:

\noindent \textbf{Situation 1}, suppose that, we have data $\left(
1,0\right) $, $\left( 0,1\right) $ as positive class and data $\left(
-1/2,0\right) $, $\left( 0,-1/2\right) $ as negative class.

\begin{equation*}
\FRAME{itbpF}{2.9706in}{2.7302in}{0in}{}{}{Figure}{\special{language
"Scientific Word";type "GRAPHIC";maintain-aspect-ratio TRUE;display
"USEDEF";valid_file "T";width 2.9706in;height 2.7302in;depth
0in;original-width 2.9274in;original-height 2.6878in;cropleft "0";croptop
"1";cropright "1";cropbottom "0";tempfilename
'NQCS3C01.wmf';tempfile-properties "XPR";}}
\end{equation*}%
Using \underline{SVM} to solve this problem, we have%
\begin{eqnarray*}
\left( w_{1},w_{2}\right) \left[ 
\begin{array}{c}
1 \\ 
0%
\end{array}%
\right] +b &=&1\text{ and} \\
\left( w_{1},w_{2}\right) \left[ 
\begin{array}{c}
0 \\ 
1%
\end{array}%
\right] +b &=&1,
\end{eqnarray*}%
which implies%
\begin{equation}
w_{1}+b=1\text{ and }w_{2}+b=1.  \tag{2.8}
\end{equation}%
\noindent For two negative points, we have%
\begin{eqnarray*}
\left( w_{1},w_{2}\right) \left[ 
\begin{array}{c}
-1/2 \\ 
0%
\end{array}%
\right] +b &=&-1,\text{ and} \\
\left( w_{1},w_{2}\right) \left[ 
\begin{array}{c}
0 \\ 
-1/2%
\end{array}%
\right] +b &=&-1,
\end{eqnarray*}%
which gives%
\begin{equation}
-\frac{w_{1}}{2}+b=-1\text{ and }-\frac{w_{2}}{2}+b=-1.  \tag{2.9}
\end{equation}%
\noindent From (2.8) and (2.9), we get $\mathbf{w}=(\frac{4}{3},\frac{4}{3})$
with $b=\frac{-1}{3},$ where $\left\Vert \mathbf{w}\right\Vert =\frac{\sqrt{%
32}}{3}$.

\noindent For \textbf{situation 2}, we consider the data $(\frac{1}{2},0)$
and $(0,\frac{1}{2})$ as positive class, data $\left( -2,0\right) $ and $%
\left( 0,-2\right) $ as negative class.%
\begin{equation*}
\FRAME{itbpF}{2.9706in}{2.7198in}{0in}{}{}{Figure}{\special{language
"Scientific Word";type "GRAPHIC";maintain-aspect-ratio TRUE;display
"USEDEF";valid_file "T";width 2.9706in;height 2.7198in;depth
0in;original-width 2.9274in;original-height 2.6775in;cropleft "0";croptop
"1";cropright "1";cropbottom "0";tempfilename
'NQCS3C02.wmf';tempfile-properties "XPR";}}
\end{equation*}
Using SVM to solve this problem, we have%
\begin{eqnarray*}
\left( w_{1},w_{2}\right) \left[ 
\begin{array}{c}
1/2 \\ 
0%
\end{array}%
\right] +b &=&1\text{ and} \\
\left( w_{1},w_{2}\right) \left[ 
\begin{array}{c}
0 \\ 
1/2%
\end{array}%
\right] +b &=&1,
\end{eqnarray*}%
which implies%
\begin{equation}
\frac{1}{2}w_{1}+b=1\text{ and }\frac{1}{2}w_{2}+b=1.  \tag{2.10}
\end{equation}%
\noindent From the negative points, we have%
\begin{eqnarray*}
\left( w_{1},w_{2}\right) \left[ 
\begin{array}{c}
-2 \\ 
0%
\end{array}%
\right] +b &=&-1\text{ and} \\
\left( w_{1},w_{2}\right) \left[ 
\begin{array}{c}
0 \\ 
-2%
\end{array}%
\right] +b &=&-1,
\end{eqnarray*}%
implies that%
\begin{equation}
-2w_{1}+b=-1\text{ and }-2w_{2}+b=-1.  \tag{2.11}
\end{equation}%
From (2.10) and (2.11), we get $\mathbf{w}=(\frac{4}{5},\frac{4}{5})$ and $b=%
\frac{3}{5}\ $with $\left\Vert \mathbf{w}\right\Vert =\frac{\sqrt{32}}{5}$.

\noindent In the next \textbf{situation 3,} we combine of this two groups of
data. Now, we have data $\left( 1/2,0\right) $, $\left( 0,1/2\right) $, $%
\left( 1,0\right) ,$ $\left( 0,1\right) $ as positive class and $\left(
-1/2,0\right) $, $\left( 0,-1/2\right) $, $\left( -2,0\right) $, $\left(
0,-2\right) $ as negative class.

\begin{equation*}
\FRAME{itbpF}{3.0649in}{2.7302in}{0in}{}{}{Figure}{\special{language
"Scientific Word";type "GRAPHIC";maintain-aspect-ratio TRUE;display
"USEDEF";valid_file "T";width 3.0649in;height 2.7302in;depth
0in;original-width 3.0208in;original-height 2.6878in;cropleft "0";croptop
"1";cropright "1";cropbottom "0";tempfilename
'NQCS3C03.wmf';tempfile-properties "XPR";}}
\end{equation*}%
Using \underline{SVM} to solve this problem, we have 
\begin{eqnarray*}
\left( w_{1},w_{2}\right) \left[ 
\begin{array}{c}
1/2 \\ 
0%
\end{array}%
\right] +b &=&1\text{ and} \\
\left( w_{1},w_{2}\right) \left[ 
\begin{array}{c}
0 \\ 
1/2%
\end{array}%
\right] +b &=&1,
\end{eqnarray*}%
which implies%
\begin{equation}
w_{1}/2+b=1\text{ and }w_{2}/2+b=1.  \tag{2.12}
\end{equation}%
\noindent For two negative points, we have%
\begin{eqnarray*}
\left( w_{1},w_{2}\right) \left[ 
\begin{array}{c}
-1/2 \\ 
0%
\end{array}%
\right] +b &=&-1\text{ and} \\
\left( w_{1},w_{2}\right) \left[ 
\begin{array}{c}
0 \\ 
-1/2%
\end{array}%
\right] +b &=&-1,
\end{eqnarray*}%
implies that%
\begin{equation}
-\frac{1}{2}w_{1}+b=-1\text{ and }-\frac{1}{2}w_{2}+b=-1.  \tag{2.13}
\end{equation}%
\noindent From (2.12) and (2.13), we obtain $\mathbf{w}=\left( 2,2\right) $
and $b=0$, where $\left\Vert \mathbf{w}\right\Vert =2\sqrt{2}$.

\noindent Now we solve the same problem for all three situations by using 
\underline{GSVM}\textbf{.}

For two positive points of \textbf{situation 1}, we have%
\begin{equation*}
\left[ 
\begin{array}{cc}
w_{11} & w_{12} \\ 
w_{21} & w_{22}%
\end{array}%
\right] \left[ 
\begin{array}{c}
1 \\ 
0%
\end{array}%
\right] +\left[ 
\begin{array}{c}
b_{1} \\ 
b_{2}%
\end{array}%
\right] =\left[ 
\begin{array}{c}
1 \\ 
1%
\end{array}%
\right]
\end{equation*}%
and%
\begin{equation*}
\left[ 
\begin{array}{cc}
w_{11} & w_{12} \\ 
w_{21} & w_{22}%
\end{array}%
\right] \left[ 
\begin{array}{c}
0 \\ 
1%
\end{array}%
\right] +\left[ 
\begin{array}{c}
b_{1} \\ 
b_{2}%
\end{array}%
\right] =\left[ 
\begin{array}{c}
1 \\ 
1%
\end{array}%
\right] ,
\end{equation*}%
which implies%
\begin{equation}
\left[ 
\begin{array}{c}
w_{11} \\ 
w_{21}%
\end{array}%
\right] +\left[ 
\begin{array}{c}
b_{1} \\ 
b_{2}%
\end{array}%
\right] =\left[ 
\begin{array}{c}
1 \\ 
1%
\end{array}%
\right] \text{ and }\left[ 
\begin{array}{c}
w_{12} \\ 
w_{22}%
\end{array}%
\right] +\left[ 
\begin{array}{c}
b_{1} \\ 
b_{2}%
\end{array}%
\right] =\left[ 
\begin{array}{c}
1 \\ 
1%
\end{array}%
\right] .  \tag{2.14}
\end{equation}

\noindent Again, for the negative points, we have%
\begin{equation*}
\left[ 
\begin{array}{cc}
w_{11} & w_{12} \\ 
w_{21} & w_{22}%
\end{array}%
\right] \left[ 
\begin{array}{c}
-1/2 \\ 
0%
\end{array}%
\right] +\left[ 
\begin{array}{c}
b_{1} \\ 
b_{2}%
\end{array}%
\right] =\left[ 
\begin{array}{c}
-1 \\ 
-1%
\end{array}%
\right]
\end{equation*}%
and%
\begin{equation*}
\left[ 
\begin{array}{cc}
w_{11} & w_{12} \\ 
w_{21} & w_{22}%
\end{array}%
\right] \left[ 
\begin{array}{c}
0 \\ 
-1/2%
\end{array}%
\right] +\left[ 
\begin{array}{c}
b_{1} \\ 
b_{2}%
\end{array}%
\right] =\left[ 
\begin{array}{c}
-1 \\ 
-1%
\end{array}%
\right] ,
\end{equation*}%
which gives%
\begin{equation}
\left[ 
\begin{array}{c}
-\frac{1}{2}w_{11} \\ 
-\frac{1}{2}w_{21}%
\end{array}%
\right] +\left[ 
\begin{array}{c}
b_{1} \\ 
b_{2}%
\end{array}%
\right] =\left[ 
\begin{array}{c}
-1 \\ 
-1%
\end{array}%
\right] \text{ and }\left[ 
\begin{array}{c}
-\frac{1}{2}w_{12} \\ 
-\frac{1}{2}w_{22}%
\end{array}%
\right] +\left[ 
\begin{array}{c}
b_{1} \\ 
b_{2}%
\end{array}%
\right] =\left[ 
\begin{array}{c}
-1 \\ 
-1%
\end{array}%
\right] .  \tag{2.15}
\end{equation}%
From (2.14) and (2.15), we get%
\begin{equation*}
W=\left[ 
\begin{array}{cc}
\frac{4}{3} & \frac{4}{3} \\ 
\frac{4}{3} & \frac{4}{3}%
\end{array}%
\right] \text{ and }B=\left[ 
\begin{array}{c}
-\frac{1}{3} \\ 
-\frac{1}{3}%
\end{array}%
\right] .
\end{equation*}%
Thus we get%
\begin{equation*}
\min_{\mathbf{w}_{i}\in W}G\left( \mathbf{w}_{i}\right) =(\frac{4\sqrt{2}}{3}%
,\frac{4\sqrt{2}}{3}).
\end{equation*}%
Hence we get $\mathbf{w}=(\frac{4}{3},\frac{4}{3})$ that minimize $G\left( 
\mathbf{w}_{i}\right) $ for $i=1,2.$

\noindent Now, for positive points of \textbf{situation 2}, we have%
\begin{equation*}
\left[ 
\begin{array}{cc}
w_{11} & w_{12} \\ 
w_{21} & w_{22}%
\end{array}%
\right] \left[ 
\begin{array}{c}
1/2 \\ 
0%
\end{array}%
\right] +\left[ 
\begin{array}{c}
b_{1} \\ 
b_{2}%
\end{array}%
\right] =\left[ 
\begin{array}{c}
1 \\ 
1%
\end{array}%
\right]
\end{equation*}%
and%
\begin{equation*}
\left[ 
\begin{array}{cc}
w_{11} & w_{12} \\ 
w_{21} & w_{22}%
\end{array}%
\right] \left[ 
\begin{array}{c}
0 \\ 
1/2%
\end{array}%
\right] +\left[ 
\begin{array}{c}
b_{1} \\ 
b_{2}%
\end{array}%
\right] =\left[ 
\begin{array}{c}
1 \\ 
1%
\end{array}%
\right] ,
\end{equation*}%
which gives%
\begin{equation*}
\left[ 
\begin{array}{c}
\frac{1}{2}w_{11} \\ 
\frac{1}{2}w_{21}%
\end{array}%
\right] +\left[ 
\begin{array}{c}
b_{1} \\ 
b_{2}%
\end{array}%
\right] =\left[ 
\begin{array}{c}
1 \\ 
1%
\end{array}%
\right] \text{ and }\left[ 
\begin{array}{c}
\frac{1}{2}w_{12} \\ 
\frac{1}{2}w_{22}%
\end{array}%
\right] +\left[ 
\begin{array}{c}
b_{1} \\ 
b_{2}%
\end{array}%
\right] =\left[ 
\begin{array}{c}
1 \\ 
1%
\end{array}%
\right] .
\end{equation*}%
\noindent For two negative points for this case, we have%
\begin{equation*}
\left[ 
\begin{array}{cc}
w_{11} & w_{12} \\ 
w_{21} & w_{22}%
\end{array}%
\right] \left[ 
\begin{array}{c}
-2 \\ 
0%
\end{array}%
\right] +\left[ 
\begin{array}{c}
b_{1} \\ 
b_{2}%
\end{array}%
\right] =\left[ 
\begin{array}{c}
-1 \\ 
-1%
\end{array}%
\right]
\end{equation*}%
and%
\begin{equation*}
\left[ 
\begin{array}{cc}
w_{11} & w_{12} \\ 
w_{21} & w_{22}%
\end{array}%
\right] \left[ 
\begin{array}{c}
0 \\ 
-2%
\end{array}%
\right] +\left[ 
\begin{array}{c}
b_{1} \\ 
b_{2}%
\end{array}%
\right] =\left[ 
\begin{array}{c}
-1 \\ 
-1%
\end{array}%
\right] ,
\end{equation*}%
which gives%
\begin{equation*}
\left[ 
\begin{array}{c}
-2w_{11} \\ 
-2w_{21}%
\end{array}%
\right] +\left[ 
\begin{array}{c}
b_{1} \\ 
b_{2}%
\end{array}%
\right] =\left[ 
\begin{array}{c}
-1 \\ 
-1%
\end{array}%
\right] \text{ and }\left[ 
\begin{array}{c}
-2w_{12} \\ 
-2w_{22}%
\end{array}%
\right] +\left[ 
\begin{array}{c}
b_{1} \\ 
b_{2}%
\end{array}%
\right] =\left[ 
\begin{array}{c}
-1 \\ 
-1%
\end{array}%
\right] .
\end{equation*}%
Thus, we obtain that%
\begin{equation*}
W=\left[ 
\begin{array}{cc}
\frac{4}{5} & \frac{4}{5} \\ 
\frac{4}{5} & \frac{4}{5}%
\end{array}%
\right] \text{ and }B=\left[ 
\begin{array}{c}
\frac{3}{5} \\ 
\frac{3}{5}%
\end{array}%
\right] .
\end{equation*}%
Thus we get%
\begin{equation*}
\min_{i\in \{1,2\}}\text{ }G\left( \mathbf{w}_{i}\right) =(\frac{4\sqrt{2}}{5%
},\frac{4\sqrt{2}}{5}).
\end{equation*}%
Hence we get $\mathbf{w}=(\frac{4}{5},\frac{4}{5})$ that minimize $G\left( 
\mathbf{w}_{i}\right) $ for $i=1,2.$

For the positive points of the combination of \textbf{situation 3}, we have%
\begin{equation*}
\left[ 
\begin{array}{cc}
w_{11} & w_{12} \\ 
w_{21} & w_{22}%
\end{array}%
\right] \left[ 
\begin{array}{c}
1/2 \\ 
0%
\end{array}%
\right] +\left[ 
\begin{array}{c}
b_{1} \\ 
b_{2}%
\end{array}%
\right] =\left[ 
\begin{array}{c}
1 \\ 
1%
\end{array}%
\right]
\end{equation*}%
and%
\begin{equation*}
\left[ 
\begin{array}{cc}
w_{11} & w_{12} \\ 
w_{21} & w_{22}%
\end{array}%
\right] \left[ 
\begin{array}{c}
0 \\ 
1/2%
\end{array}%
\right] +\left[ 
\begin{array}{c}
b_{1} \\ 
b_{2}%
\end{array}%
\right] =\left[ 
\begin{array}{c}
1 \\ 
1%
\end{array}%
\right] ,
\end{equation*}%
which gives%
\begin{equation*}
\left[ 
\begin{array}{c}
\frac{1}{2}w_{11} \\ 
\frac{1}{2}w_{21}%
\end{array}%
\right] +\left[ 
\begin{array}{c}
b_{1} \\ 
b_{2}%
\end{array}%
\right] =\left[ 
\begin{array}{c}
1 \\ 
1%
\end{array}%
\right] \text{ and }\left[ 
\begin{array}{c}
\frac{1}{2}w_{12} \\ 
\frac{1}{2}w_{22}%
\end{array}%
\right] +\left[ 
\begin{array}{c}
b_{1} \\ 
b_{2}%
\end{array}%
\right] =\left[ 
\begin{array}{c}
1 \\ 
1%
\end{array}%
\right] .
\end{equation*}%
For two negative points for this case, we have%
\begin{equation*}
\left[ 
\begin{array}{cc}
w_{11} & w_{12} \\ 
w_{21} & w_{22}%
\end{array}%
\right] \left[ 
\begin{array}{c}
-\frac{1}{2} \\ 
0%
\end{array}%
\right] +\left[ 
\begin{array}{c}
b_{1} \\ 
b_{2}%
\end{array}%
\right] =\left[ 
\begin{array}{c}
-1 \\ 
-1%
\end{array}%
\right]
\end{equation*}%
and%
\begin{equation*}
\left[ 
\begin{array}{cc}
w_{11} & w_{12} \\ 
w_{21} & w_{22}%
\end{array}%
\right] \left[ 
\begin{array}{c}
0 \\ 
-\frac{1}{2}%
\end{array}%
\right] +\left[ 
\begin{array}{c}
b_{1} \\ 
b_{2}%
\end{array}%
\right] =\left[ 
\begin{array}{c}
-1 \\ 
-1%
\end{array}%
\right] ,
\end{equation*}%
which gives%
\begin{equation*}
\left[ 
\begin{array}{c}
-\frac{1}{2}w_{11} \\ 
-\frac{1}{2}w_{21}%
\end{array}%
\right] +\left[ 
\begin{array}{c}
b_{1} \\ 
b_{2}%
\end{array}%
\right] =\left[ 
\begin{array}{c}
-1 \\ 
-1%
\end{array}%
\right] \text{ and }\left[ 
\begin{array}{c}
-\frac{1}{2}w_{12} \\ 
-\frac{1}{2}w_{22}%
\end{array}%
\right] +\left[ 
\begin{array}{c}
b_{1} \\ 
b_{2}%
\end{array}%
\right] =\left[ 
\begin{array}{c}
-1 \\ 
-1%
\end{array}%
\right] .
\end{equation*}%
From this, we obtain that%
\begin{equation*}
W=\left[ 
\begin{array}{cc}
2 & 2 \\ 
2 & 2%
\end{array}%
\right] \text{ and }B=\left[ 
\begin{array}{c}
0 \\ 
0%
\end{array}%
\right] .
\end{equation*}%
Thus we get%
\begin{equation*}
\min_{i\in \{1,2\}}\text{ }G\left( \mathbf{w}_{i}\right) =(2\sqrt{2},2\sqrt{2%
}).
\end{equation*}%
Hence we get $\mathbf{w}=(2,2)$ that minimize $G\left( \mathbf{w}_{i}\right) 
$ for $i=1,2.$ $\square $\bigskip

\noindent \textbf{Proposition 2.4.} \ \ Let $G:%
\mathbb{R}
^{n}\rightarrow 
\mathbb{R}
_{+}^{n}$ be a differentiable operator. An element $\mathbf{w}^{\ast }\in 
\mathbb{R}
^{n}$ minimize $G$ if and only if $G^{\prime }\left( \mathbf{w}^{\ast
}\right) =0,$ that is, $\mathbf{w}^{\ast }\in 
\mathbb{R}
^{n}$ solves GSVM if and only if $G^{\prime }\left( \mathbf{w}^{\ast
}\right) =0.$

\noindent \textbf{Proof.} \ \ Let $G^{\prime }\left( \mathbf{w}^{\ast
}\right) =0,$ then for all $\mathbf{v}\in 
\mathbb{R}
^{n},$%
\begin{equation*}
<\eta G^{\prime }\left( \mathbf{w}^{\ast }\right) ,\mathbf{v}-\mathbf{w}%
^{\ast }>\text{ }=\text{ }<0,\mathbf{v}-\mathbf{w}^{\ast }>\text{ }=\text{ }%
0.
\end{equation*}%
Consequently, the inequality%
\begin{equation*}
<\eta G^{\prime }\left( \mathbf{w}^{\ast }\right) ,\mathbf{v}-\mathbf{w}%
^{\ast }>\text{ }=\text{ }<0,\mathbf{v}-\mathbf{w}^{\ast }>\text{ }\geq 0
\end{equation*}%
holds for all $\mathbf{v}\in 
\mathbb{R}
^{n}.$ Hence $\mathbf{w}^{\ast }\in 
\mathbb{R}
^{n}$ solves problem of GSVM.

\noindent Conversely, assume that $\mathbf{w}^{\ast }\in $ $%
\mathbb{R}
^{n}$ satisfies%
\begin{equation*}
<\eta G^{\prime }\left( \mathbf{w}^{\ast }\right) ,\mathbf{v}-\mathbf{w}%
^{\ast }>\text{ }\geq 0\text{ \ \ }\forall \text{ }\mathbf{v}\in 
\mathbb{R}
^{n}.
\end{equation*}%
Take $\mathbf{v}=\mathbf{w}^{\ast }-G^{\prime }\left( \mathbf{w}^{\ast
}\right) $ in the above inequality implies that%
\begin{equation*}
<\eta G^{\prime }\left( \mathbf{w}^{\ast }\right) ,-G^{\prime }\left( 
\mathbf{w}^{\ast }\right) >\text{ }\geq \text{ }0,
\end{equation*}%
which further implies%
\begin{equation*}
-\eta ||G^{\prime }(\mathbf{w}^{\ast })||^{2}\text{ }\geq \text{ }0.
\end{equation*}%
Since $\eta >0,$ so we get $G^{\prime }(\mathbf{w}^{\ast })=0.$ $\square $%
\bigskip

\noindent \textbf{Definition 2.5.} \ \ Let $K$ be a closed and convex subset
of $%
\mathbb{R}
^{n}$. Then, for every point $\mathbf{x}\in 
\mathbb{R}
^{n}$, there exists a unique nearest point in $K$, denoted by $P_{K}\left( 
\mathbf{x}\right) $, such that $\left\Vert \mathbf{x}-P_{K}\left( \mathbf{x}%
\right) \right\Vert \leq \left\Vert \mathbf{x}-\mathbf{y}\right\Vert $ for
all $\mathbf{y}\in K$ and also note that $P_{K}\left( \mathbf{x}\right) =%
\mathbf{x}$ if $\mathbf{x}\in K$. $P_{K}$ is called the metric projection of 
$%
\mathbb{R}
^{n}$ onto $K$. It is well known that $P_{K}:%
\mathbb{R}
^{n}\rightarrow K$ is characterized by the properties:

\begin{enumerate}
\item[(i)] $P_{K}\left( \mathbf{x}\right) =\mathbf{z}$ for $\mathbf{x}\in 
\mathbb{R}
^{n}$ if and only if $<\mathbf{z},\mathbf{y}-\mathbf{z}>$ $\geq $\ $<\mathbf{%
x},\mathbf{y}-\mathbf{z}>$ for all $\mathbf{y}\in 
\mathbb{R}
^{n}$;

\item[(ii)] For every $\mathbf{x,y}\in 
\mathbb{R}
^{n}$, $\left\Vert P_{K}\left( \mathbf{x}\right) -P_{K}\left( \mathbf{y}%
\right) \right\Vert ^{2}$ $\leq $ $<\mathbf{x}-\mathbf{y},P_{K}\left( 
\mathbf{x}\right) -P_{K}\left( \mathbf{y}\right) >$;

\item[(iii)] $\left\Vert P_{K}\left( \mathbf{x}\right) -P_{K}\left( \mathbf{y%
}\right) \right\Vert $ $\leq $ $\left\Vert \mathbf{x}-\mathbf{y}\right\Vert $%
, for every $\mathbf{x,y}\in 
\mathbb{R}
^{n},$ that is, $P_{K}$ is nonexpansive map.
\end{enumerate}

\noindent \textbf{Proposition 2.6.}\ \ \ Let $G:%
\mathbb{R}
^{n}\rightarrow 
\mathbb{R}
_{+}^{n}$ be a differentiable operator. An element $\mathbf{w}^{\ast }\in 
\mathbb{R}
^{n}$ minimize mapping $G$ defined in (2.3)\ if and only if $\mathbf{w}%
^{\ast }$ is the fixed point of map%
\begin{equation*}
P_{%
\mathbb{R}
_{+}^{n}}\left( I-\rho G^{\prime }\right) :%
\mathbb{R}
^{n}\rightarrow 
\mathbb{R}
_{+}^{n}\text{ for any }\rho >0.
\end{equation*}%
that is,%
\begin{eqnarray*}
\mathbf{w}^{\ast } &=&P_{%
\mathbb{R}
_{+}^{n}}\left( I-\rho G^{\prime }\right) (\mathbf{w}^{\ast }) \\
&=&P_{%
\mathbb{R}
_{+}^{n}}\left( \mathbf{w}^{\ast }-\rho G^{\prime }\left( \mathbf{w}^{\ast
}\right) \right) ,
\end{eqnarray*}%
where $P_{%
\mathbb{R}
_{+}^{n}}$ is a projection map from $%
\mathbb{R}
^{n}$ to $%
\mathbb{R}
_{+}^{n}.$

\noindent \textbf{Proof. \ \ }Suppose $\mathbf{w}^{\ast }\in 
\mathbb{R}
_{+}^{n}$ is solution of $GSVM$ then for $\eta >0,$ we have%
\begin{equation*}
<\eta G^{\prime }\left( \mathbf{w}^{\ast }\right) ,\mathbf{w}-\mathbf{w}%
^{\ast }>\text{ }\geq \text{ }0\text{ for all }\mathbf{w}\in 
\mathbb{R}
^{n}.
\end{equation*}%
Adding $<\mathbf{w}^{\ast },\mathbf{w}-\mathbf{w}^{\ast }>$ on both sides,
we get%
\begin{equation*}
<\mathbf{w}^{\ast },\mathbf{w}-\mathbf{w}^{\ast }>\text{ }+\text{ }<\eta
G^{\prime }\left( \mathbf{w}^{\ast }\right) ,\mathbf{w}-\mathbf{w}^{\ast }>%
\text{ }\geq \text{ }<\mathbf{w}^{\ast },\mathbf{w}-\mathbf{w}^{\ast }>\ 
\text{for all }\mathbf{w}\in 
\mathbb{R}
^{n},
\end{equation*}%
which further implies that%
\begin{equation*}
<\mathbf{w}^{\ast },\mathbf{w}-\mathbf{w}^{\ast }>\ \geq \text{ }<\mathbf{w}%
^{\ast }-\eta G^{\prime }\left( \mathbf{w}^{\ast }\right) ,\mathbf{w}-%
\mathbf{w}^{\ast }>\ \text{for all }\mathbf{w}\in 
\mathbb{R}
^{n},
\end{equation*}%
which is possible only if $\mathbf{w}^{\ast }=P_{%
\mathbb{R}
_{+}^{n}}\left( \mathbf{w}^{\ast }-\rho G^{\prime }\left( \mathbf{w}^{\ast
}\right) \right) ,$ that is, $\mathbf{w}^{\ast }$ is the fixed point of $%
G^{\prime }.$

Conversely, let $\mathbf{w}^{\ast }=P_{%
\mathbb{R}
_{+}^{n}}\left( \mathbf{w}^{\ast }-\rho G^{\prime }\left( \mathbf{w}^{\ast
}\right) \right) ,$ then we have%
\begin{equation*}
<\mathbf{w}^{\ast },\mathbf{w}-\mathbf{w}^{\ast }>\ \geq \text{ }<\mathbf{w}%
^{\ast }-\eta G^{\prime }\left( \mathbf{w}^{\ast }\right) ,\mathbf{w}-%
\mathbf{w}^{\ast }>\ \text{for all }\mathbf{w}\in 
\mathbb{R}
^{n},
\end{equation*}%
which implies%
\begin{equation*}
<\mathbf{w}^{\ast },\mathbf{w}-\mathbf{w}^{\ast }>-<\mathbf{w}^{\ast }-\eta
G^{\prime }\left( \mathbf{w}^{\ast }\right) ,\mathbf{w}-\mathbf{w}^{\ast }>\
\geq \ 0\ \text{for all }\mathbf{w}\in 
\mathbb{R}
^{n},
\end{equation*}%
and so%
\begin{equation*}
<\eta G^{\prime }\left( \mathbf{w}^{\ast }\right) ,\mathbf{w}-\mathbf{w}%
^{\ast }>\ \geq \text{ }0\text{\ for all }\mathbf{w}\in 
\mathbb{R}
^{n}.
\end{equation*}%
Thus $\mathbf{w}^{\ast }\in 
\mathbb{R}
_{+}^{n}$ is the solution of GSVM. $\square $\bigskip

\noindent \textbf{Definition} \textbf{2.7.}\ \ \ A map $G:%
\mathbb{R}
^{n}\rightarrow 
\mathbb{R}
^{n}$ is said to be

\noindent (I) $L$-Lipschitz if for every $L>0$,%
\begin{equation*}
\left\Vert G\left( \mathbf{x}\right) -G\left( \mathbf{y}\right) \right\Vert
\leq L\left\Vert \mathbf{x}-\mathbf{y}\right\Vert \text{ for all }\mathbf{x},%
\mathbf{y}\in 
\mathbb{R}
^{n}.
\end{equation*}

\noindent (II) monotone if%
\begin{equation*}
<G\left( \mathbf{x}\right) -G\left( \mathbf{y}\right) ,\mathbf{x}-\mathbf{y}>%
\text{ }\geq \ 0\text{ for all }\mathbf{x},\mathbf{y}\in 
\mathbb{R}
^{n}.
\end{equation*}

\noindent (III) strictly monotone if%
\begin{equation*}
<G\left( \mathbf{x}\right) -G\left( \mathbf{y}\right) ,\mathbf{x}-\mathbf{y}>%
\text{ }>\ 0\text{ for all }\mathbf{x},\mathbf{y}\in 
\mathbb{R}
^{n}\text{ with }\mathbf{x}\neq \mathbf{y}.
\end{equation*}%
\bigskip \noindent (IV) $\alpha $-strongly monotone if%
\begin{equation*}
<G\left( \mathbf{x}\right) -G\left( \mathbf{y}\right) ,\mathbf{x}-\mathbf{y}>%
\text{ }\geq \ \alpha \left\Vert \mathbf{x}-\mathbf{y}\right\Vert ^{2}\text{
for all }\mathbf{x},\mathbf{y}\in 
\mathbb{R}
^{n}.
\end{equation*}%
Note that, every $\alpha $-strongly monotone\ map $G:%
\mathbb{R}
^{n}\rightarrow 
\mathbb{R}
^{n}$ is strictly monotone and every strictly monotone map is monotone.%
\newline

\noindent \textbf{Example 2.8.}\ \ Let $G:%
\mathbb{R}
^{n}\rightarrow 
\mathbb{R}
^{n}$ be a mapping defined as%
\begin{equation*}
G\left( \mathbf{w}_{i}\right) =\alpha \mathbf{w}_{i}+\beta ,
\end{equation*}
where $\alpha $ is any non negative scalar and $\beta $ is any real number.
Then $G$ is Lipschitz continuous with Lipschitz constant $L=\alpha $.

Also, for any $\mathbf{x,y}\in 
\mathbb{R}
^{n},$%
\begin{equation*}
<G\left( \mathbf{x}\right) -G\left( \mathbf{y}\right) ,\mathbf{x}-\mathbf{y}>%
\text{ }=\text{ }\alpha \left\Vert \mathbf{x}-\mathbf{y}\right\Vert ^{2}
\end{equation*}%
which show that $G$ is $\alpha $-strongly monotone. $\square $\bigskip

\noindent \textbf{Theorem 2.9.\ \ \ }Let $K\subseteq 
\mathbb{R}
^{n}$ be closed and convex and $G^{\prime }:%
\mathbb{R}
^{n}\rightarrow K$ is strictly monotone. If there exists a $\mathbf{w}^{\ast
}\in K$ which is the solution of $GSVM,$ then $\mathbf{w}^{\ast }$ is unique
in $K$.

\noindent \textbf{Proof. \ \ }Suppose that $\mathbf{w}_{1}^{\ast },\mathbf{w}%
_{2}^{\ast }\in K$ with $\mathbf{w}_{1}^{\ast }\neq \mathbf{w}_{2}^{\ast }$
be the two solutions of $GSVM,$ then we have%
\begin{equation}
<\eta G^{\prime }\left( \mathbf{w}_{1}^{\ast }\right) ,\mathbf{w}-\mathbf{w}%
_{1}^{\ast }>\text{ }\geq \text{ }0\text{ for all }\mathbf{w}\in 
\mathbb{R}
^{n}  \tag{2.16}
\end{equation}%
and%
\begin{equation}
<\eta G^{\prime }\left( \mathbf{w}_{2}^{\ast }\right) ,\mathbf{w}-\mathbf{w}%
_{2}^{\ast }>\text{ }\geq \text{ }0\text{ for all }\mathbf{w}\in 
\mathbb{R}
^{n},  \tag{2.17}
\end{equation}%
where $\eta >0$. Putting $\mathbf{w}=\mathbf{w}_{2}^{\ast }$ in (2.16) and $%
\mathbf{w}=\mathbf{w}_{1}^{\ast }$ in (2.17), we get%
\begin{equation}
<\eta G^{\prime }\left( \mathbf{w}_{1}^{\ast }\right) ,\mathbf{w}_{2}^{\ast
}-\mathbf{w}_{1}^{\ast }>\text{ }\geq \text{ }0  \tag{2.18}
\end{equation}%
and%
\begin{equation}
<\eta G^{\prime }\left( \mathbf{w}_{2}^{\ast }\right) ,\mathbf{w}_{1}^{\ast
}-\mathbf{w}_{2}^{\ast }>\text{ }\geq \text{ }0.  \tag{2.19}
\end{equation}%
Eq. (2.18) can be further write as%
\begin{equation}
<-\eta G^{\prime }\left( \mathbf{w}_{1}^{\ast }\right) ,\mathbf{w}_{1}^{\ast
}-\mathbf{w}_{2}^{\ast }>\text{ }\geq \text{ }0.  \tag{2.20}
\end{equation}%
Adding (2.19) and (2.20) implies that%
\begin{equation*}
<\eta G^{\prime }\left( \mathbf{w}_{2}^{\ast }\right) -\eta G^{\prime
}\left( \mathbf{w}_{1}^{\ast }\right) ,\mathbf{w}_{1}^{\ast }-\mathbf{w}%
_{2}^{\ast }>\text{ }\geq \text{ }0
\end{equation*}%
which implies%
\begin{equation*}
\eta <G^{\prime }\left( \mathbf{w}_{1}^{\ast }\right) -G^{\prime }\left( 
\mathbf{w}_{2}^{\ast }\right) ,\mathbf{w}_{1}^{\ast }-\mathbf{w}_{2}^{\ast }>%
\text{ }\leq \text{ }0
\end{equation*}%
or%
\begin{equation}
<G^{\prime }\left( \mathbf{w}_{1}^{\ast }\right) -G^{\prime }\left( \mathbf{w%
}_{2}^{\ast }\right) ,\mathbf{w}_{1}^{\ast }-\mathbf{w}_{2}^{\ast }>\text{ }%
\leq \text{ }0.  \tag{2.21}
\end{equation}%
Since $G^{\prime }$ is strictly monotone, so we must have%
\begin{equation*}
<G^{\prime }\left( \mathbf{w}_{1}^{\ast }\right) -G^{\prime }\left( \mathbf{w%
}_{2}^{\ast }\right) ,\mathbf{w}_{1}^{\ast }-\mathbf{w}_{2}^{\ast }>\text{ }>%
\text{ }0,
\end{equation*}%
which contradicts (2.21). Thus $\mathbf{w}_{1}^{\ast }=\mathbf{w}_{2}^{\ast
} $. $\square \medskip $

\noindent \textbf{Theorem 2.10.}\ \ \ Let $K\subseteq 
\mathbb{R}
^{n}$ be closed and convex. If the map $G^{\prime }:%
\mathbb{R}
^{n}\rightarrow K$ is $L$-Lipchitz and $\alpha $-strongly monotone then
there exists a unique $\mathbf{w}^{\ast }\in K$ which is the solution of $%
GSVM$.

\noindent \textbf{Proof.}$\mathbf{\ }$\textbf{\ Uniqueness}:

\noindent Suppose that $\mathbf{w}_{1}^{\ast },\mathbf{w}_{2}^{\ast }\in K$
be the two solutions of $GSVM,$ then for $\eta >0,$ we have%
\begin{equation}
<\eta G^{\prime }\left( \mathbf{w}_{1}^{\ast }\right) ,\mathbf{w}-\mathbf{w}%
_{1}^{\ast }>\text{ }\geq \text{ }0\text{ for all }\mathbf{w}\in 
\mathbb{R}
^{n}  \tag{2.22}
\end{equation}%
and%
\begin{equation}
<\eta G^{\prime }\left( \mathbf{w}_{2}^{\ast }\right) ,\mathbf{w}-\mathbf{w}%
_{2}^{\ast }>\text{ }\geq \text{ }0\text{ for all }\mathbf{w}\in 
\mathbb{R}
^{n}.  \tag{2.23}
\end{equation}%
Putting $\mathbf{w}=\mathbf{w}_{2}^{\ast }$ in (2.22) and $\mathbf{w}=%
\mathbf{w}_{1}^{\ast }$ in (2.23), we get%
\begin{equation}
<\eta G^{\prime }\left( \mathbf{w}_{1}^{\ast }\right) ,\mathbf{w}_{2}^{\ast
}-\mathbf{w}_{1}^{\ast }>\text{ }\geq \text{ }0  \tag{2.24}
\end{equation}%
and%
\begin{equation}
<\eta G^{\prime }\left( \mathbf{w}_{2}^{\ast }\right) ,\mathbf{w}_{1}^{\ast
}-\mathbf{w}_{2}^{\ast }>\text{ }\geq \text{ }0.  \tag{2.25}
\end{equation}%
Eq. (2.24) can be further write as%
\begin{equation}
<-\eta G^{\prime }\left( \mathbf{w}_{1}^{\ast }\right) ,\mathbf{w}_{1}^{\ast
}-\mathbf{w}_{2}^{\ast }>\text{ }\geq \text{ }0.  \tag{2.26}
\end{equation}%
Adding (2.25) and (2.26) implies that%
\begin{equation*}
<\eta G^{\prime }\left( \mathbf{w}_{2}^{\ast }\right) -\eta G^{\prime
}\left( \mathbf{w}_{1}^{\ast }\right) ,\mathbf{w}_{1}^{\ast }-\mathbf{w}%
_{2}^{\ast }>\text{ }\geq \text{ }0
\end{equation*}%
which implies%
\begin{equation}
\eta <G^{\prime }\left( \mathbf{w}_{1}^{\ast }\right) -G^{\prime }\left( 
\mathbf{w}_{2}^{\ast }\right) ,\mathbf{w}_{1}^{\ast }-\mathbf{w}_{2}^{\ast }>%
\text{ }\leq \text{ }0.  \tag{2.27}
\end{equation}%
Since $G^{\prime }$ is $\alpha $-strongly monotone, so we have%
\begin{eqnarray*}
\alpha \eta \left\Vert \mathbf{w}_{1}^{\ast }-\mathbf{w}_{2}^{\ast
}\right\Vert ^{2}\text{ } &\leq &\text{ }\eta <G^{\prime }\left( \mathbf{w}%
_{1}^{\ast }\right) -G^{\prime }\left( \mathbf{w}_{2}^{\ast }\right) ,%
\mathbf{w}_{1}^{\ast }-\mathbf{w}_{2}^{\ast }>\text{ } \\
&\leq &\text{ }0,
\end{eqnarray*}%
which implies that%
\begin{equation*}
\alpha \eta \left\Vert \mathbf{w}_{1}^{\ast }-\mathbf{w}_{2}^{\ast
}\right\Vert ^{2}\text{ }\leq \text{ }0\text{.}
\end{equation*}%
Since $\alpha \eta >0,$ so we must have $\left\Vert \mathbf{w}_{1}^{\ast }-%
\mathbf{w}_{2}^{\ast }\right\Vert =0$ and hence $\mathbf{w}_{1}^{\ast }=%
\mathbf{w}_{2}^{\ast }$.

\noindent \textbf{Existence}:

\noindent As we know that if $\mathbf{w}^{\ast }\in 
\mathbb{R}
_{+}^{n}$ is solution of $GSVM$ then for $\eta >0,$ we have%
\begin{equation*}
<\eta G^{\prime }\left( \mathbf{w}^{\ast }\right) ,\mathbf{w}-\mathbf{w}%
^{\ast }>\text{ }\geq 0\text{ for all }\mathbf{w}\in 
\mathbb{R}
^{n}
\end{equation*}%
if and only if%
\begin{equation*}
\begin{array}{ll}
\mathbf{w}^{\ast } & =P_{%
\mathbb{R}
_{+}^{n}}(\mathbf{w}^{\ast }-\rho G^{\prime }(\mathbf{w}^{\ast })) \\ 
& \equiv F\left( \mathbf{w}^{\ast }\right) \text{ (say).}%
\end{array}%
\end{equation*}%
Now for any $\mathbf{w}_{1}^{\ast },\mathbf{w}_{2}^{\ast }\in 
\mathbb{R}
_{+}^{n}$, we have%
\begin{equation*}
\begin{array}{l}
\left\Vert F\left( \mathbf{w}_{1}^{\ast }\right) -F\left( \mathbf{w}%
_{2}^{\ast }\right) \right\Vert ^{2} \\ 
=\left\Vert P_{%
\mathbb{R}
_{+}^{n}}(\mathbf{w}_{1}^{\ast }-\rho G^{\prime }(\mathbf{w}_{1}^{\ast
}))-P_{%
\mathbb{R}
_{+}^{n}}(\mathbf{w}_{2}^{\ast }-\rho G^{\prime }(\mathbf{w}_{2}^{\ast
}))\right\Vert ^{2} \\ 
\leq \left\Vert (\mathbf{w}_{1}^{\ast }-\rho G^{\prime }(\mathbf{w}%
_{1}^{\ast }))-(\mathbf{w}_{2}^{\ast }-\rho G^{\prime }(\mathbf{w}_{2}^{\ast
}))\right\Vert ^{2}\text{ (as }P_{%
\mathbb{R}
_{+}^{n}}\text{ is nonexpansive)} \\ 
=\left\Vert (\mathbf{w}_{1}^{\ast }-\mathbf{w}_{2}^{\ast })-\rho \lbrack
G^{\prime }(\mathbf{w}_{1}^{\ast })-G^{\prime }(\mathbf{w}_{2}^{\ast
})]\right\Vert ^{2} \\ 
=\text{ }<(\mathbf{w}_{1}^{\ast }{\small -}\mathbf{w}_{2}^{\ast })-\rho
\lbrack G^{\prime }(\mathbf{w}_{1}^{\ast }){\small -}G^{\prime }(\mathbf{w}%
_{2}^{\ast })],(\mathbf{w}_{1}^{\ast }{\small -}\mathbf{w}_{2}^{\ast })-\rho
\lbrack G^{\prime }(\mathbf{w}_{1}^{\ast }){\small -}G^{\prime }(\mathbf{w}%
_{2}^{\ast })]> \\ 
=\left\Vert \mathbf{w}_{1}^{\ast }{\small -}\mathbf{w}_{2}^{\ast
}\right\Vert ^{2}-2\rho <\mathbf{w}_{1}^{\ast }{\small -}\mathbf{w}%
_{2}^{\ast },G^{\prime }(\mathbf{w}_{1}^{\ast }){\small -}G^{\prime }(%
\mathbf{w}_{2}^{\ast })>+\rho ^{2}\left\Vert G^{\prime }(\mathbf{w}%
_{1}^{\ast }){\small -}G^{\prime }(\mathbf{w}_{2}^{\ast })\right\Vert ^{2}.%
\end{array}%
\end{equation*}%
Now since $G^{\prime }$ is $L$-Lipchitz and $\alpha $-strongly monotone, so
we get%
\begin{eqnarray*}
\left\Vert F\left( \mathbf{w}_{1}^{\ast }\right) -F\left( \mathbf{w}%
_{2}^{\ast }\right) \right\Vert ^{2} &\leq &\left\Vert \mathbf{w}_{1}^{\ast
}-\mathbf{w}_{2}^{\ast }\right\Vert ^{2}-2\alpha \rho \left\Vert \mathbf{w}%
_{1}^{\ast }-\mathbf{w}_{2}^{\ast }\right\Vert ^{2} \\
&&+\rho ^{2}L^{2}\left\Vert \mathbf{w}_{1}^{\ast }-\mathbf{w}_{2}^{\ast
}\right\Vert ^{2} \\
&=&(1+\rho ^{2}L^{2}-2\rho \alpha )\left\Vert \mathbf{w}_{1}^{\ast }-\mathbf{%
w}_{2}^{\ast }\right\Vert ^{2},
\end{eqnarray*}%
that is,%
\begin{equation}
\left\Vert F\left( \mathbf{w}_{1}^{\ast }\right) -F\left( \mathbf{w}%
_{2}^{\ast }\right) \right\Vert \leq \theta \left\Vert \mathbf{w}_{1}^{\ast
}-\mathbf{w}_{2}^{\ast }\right\Vert ,  \tag{2.30}
\end{equation}%
where $\theta =\sqrt{1+\rho ^{2}L^{2}-2\rho \alpha }.$ Since $\rho >0$, so
that when $\rho \in (0,\frac{2\alpha }{L^{2}}),$ then we get $\theta \in
\lbrack 0,1).$ Now, by using Principle of Banach contraction, we obtain the
fixed point of map $F,$ that is, there exists a unique $\mathbf{w}^{\ast
}\in 
\mathbb{R}
_{+}^{n}$ such that%
\begin{eqnarray*}
F\left( \mathbf{w}^{\ast }\right) &=&P_{%
\mathbb{R}
_{+}^{n}}(\mathbf{w}^{\ast }-\rho G^{\prime }(\mathbf{w}^{\ast })) \\
&=&\mathbf{w}^{\ast }.
\end{eqnarray*}%
Hence $\mathbf{w}^{\ast }\in 
\mathbb{R}
_{+}^{n}$ is the solution of GSVM. $\square $\bigskip

\noindent \textbf{Example 2.11.}\ \ \ Let us take the group of data of
positive class $\left( \alpha _{1},\alpha _{2},...,\alpha _{n-1},0\right) ,$ 
$(\alpha _{1},\alpha _{2},...,\alpha _{n-2},0,\alpha _{n}),$ $...,$ $\left(
0,\alpha _{2},\alpha _{3},...,\alpha _{n}\right) $ and negative class $%
\left( k\alpha _{1},k\alpha _{2},...,k\alpha _{n-1},0\right) ,$ $(k\alpha
_{1},k\alpha _{2},...,k\alpha _{n-2},0,k\alpha _{n}),$ $...,$ $\left(
0,k\alpha _{2},k\alpha _{3},...,k\alpha _{n}\right) $ for $n\geq 2,$ where
each $\alpha _{i}\neq 0$ for $i\in 
\mathbb{N}
$ and $k\neq 1.$

\noindent A map $G:%
\mathbb{R}
^{n}\rightarrow 
\mathbb{R}
_{+}^{n}$ be given as%
\begin{equation*}
G\left( \mathbf{w}_{i}\right) =\left( \left\Vert \mathbf{w}_{i}\right\Vert
,\left\Vert \mathbf{w}_{i}\right\Vert ,...,\left\Vert \mathbf{w}%
_{i}\right\Vert \right) \quad \text{for}\quad i=1,2,...,n,
\end{equation*}%
where $\mathbf{w}_{i}$ are the row of $W_{n\times n}$ for $i=1,2,...,n.$
Then we have%
\begin{equation*}
G^{\prime }\left( \mathbf{w}_{i}\right) =\frac{1}{\left\Vert \mathbf{w}%
_{i}\right\Vert }\mathbf{w}_{i}\quad \text{for}\quad i=1,2,...,n.
\end{equation*}%
Now from the given data, we get%
\begin{equation*}
W=\frac{2}{\left( n-1\right) \left( 1-k\right) }\left[ 
\begin{array}{cccc}
\frac{1}{\alpha _{1}} & \frac{1}{\alpha _{2}} & \cdot \cdot \cdot & \frac{1}{%
\alpha _{n}} \\ 
\frac{1}{\alpha _{1}} & \frac{1}{\alpha _{2}} & \cdot \cdot \cdot & \frac{1}{%
\alpha _{n}} \\ 
\begin{array}{c}
\cdot \\ 
\cdot%
\end{array}
& 
\begin{array}{c}
\cdot \\ 
\cdot%
\end{array}
& 
\begin{array}{c}
\cdot \\ 
\cdot%
\end{array}
& 
\begin{array}{c}
\cdot \\ 
\cdot%
\end{array}
\\ 
\frac{1}{\alpha _{1}} & \frac{1}{\alpha _{2}} & \cdot \cdot \cdot & \frac{1}{%
\alpha _{n}}%
\end{array}%
\right]
\end{equation*}%
and so we have%
\begin{equation*}
G\left( \mathbf{w}_{i}\right) =\frac{2}{\left( n-1\right) \left( 1-k\right) }%
\sqrt{\frac{1}{\alpha _{1}^{2}}+\frac{1}{\alpha _{2}^{2}}+...+\frac{1}{%
\alpha _{n}^{2}}}(1,1,...,1)\text{ for }i=1,2,...,n
\end{equation*}%
and%
\begin{equation*}
G^{\prime }\left( \mathbf{w}_{i}\right) =\frac{1}{\sqrt{\frac{1}{\alpha
_{1}^{2}}+\frac{1}{\alpha _{2}^{2}}+...+\frac{1}{\alpha _{n}^{2}}}}(\frac{1}{%
\alpha _{1}},\frac{1}{\alpha _{2}},...,\frac{1}{\alpha _{n}})\text{ for }%
i=1,2,...,n.
\end{equation*}%
Note that, for any $\mathbf{w}_{1},\mathbf{w}_{2}\in W,$%
\begin{equation*}
\left\Vert G^{\prime }\left( \mathbf{w}_{1}\right) -G^{\prime }\left( 
\mathbf{w}_{2}\right) \right\Vert =0=L\left\Vert \mathbf{w}_{1}-\mathbf{w}%
_{2}\right\Vert
\end{equation*}%
is satisfied where $L$ is any nonnegative real number. Also%
\begin{equation*}
<G^{\prime }\left( \mathbf{w}_{1}\right) -G^{\prime }\left( \mathbf{w}%
_{2}\right) ,\mathbf{w}_{1}-\mathbf{w}_{2}>\text{ }\geq \text{ }0
\end{equation*}%
is satisfied which show that $G^{\prime }$ is monotone operator. Moreover, $%
\mathbf{w}=\frac{2}{\left( n-1\right) \left( 1-k\right) }(\frac{1}{\alpha
_{1}},\frac{1}{\alpha _{2}},...,\frac{1}{\alpha _{n}})$ is the solution of
GSVM with $\left\Vert \mathbf{w}\right\Vert =\frac{2}{\left( n-1\right)
\left( 1-k\right) }\sqrt{\frac{1}{\alpha _{1}^{2}}+\frac{1}{\alpha _{2}^{2}}%
+...+\frac{1}{\alpha _{n}^{2}}}.$ $\square $\newline

\noindent \textbf{Example 2.12.}\ \ \ Let us take the group of data of
positive class $\left( \alpha _{1},\alpha _{2},...,\alpha
_{m},0,0...,0\right) ,$ $(0,\alpha _{2},\alpha _{3},...,\alpha
_{m+1},0,0,...,0),$ $...,$ $\left( \alpha _{1},\alpha _{2},...,\alpha
_{m-1},0,0,...,0,\alpha _{n}\right) $ and negative class $\left( \kappa
\alpha _{1},\kappa \alpha _{2},...,\kappa \alpha _{m},0,0...,0\right) ,$ $%
(0,\kappa \alpha _{2},\kappa \alpha _{3},...,\kappa \alpha
_{m+1},0,0,...,0), $ $...,$ $\left( \kappa \alpha _{1},\kappa \alpha
_{2},...,\kappa \alpha _{m-1},0,0,...,0,\kappa \alpha _{n}\right) $ for $%
n>m\geq 1,$ where each $\alpha _{i}\neq 0$ for $i\in 
\mathbb{N}
$ and $\kappa \neq 1.$

\noindent A map $G:%
\mathbb{R}
^{n}\rightarrow 
\mathbb{R}
_{+}^{n}$ be given as%
\begin{equation*}
G\left( \mathbf{w}_{i}\right) =\left( \left\Vert \mathbf{w}_{i}\right\Vert
,\left\Vert \mathbf{w}_{i}\right\Vert ,...,\left\Vert \mathbf{w}%
_{i}\right\Vert \right) \quad \text{for}\quad i=1,2,...,n,
\end{equation*}%
where $\mathbf{w}_{i}$ are the row of $W_{n\times n}$ for $i=1,2,...,n.$
Then we have%
\begin{equation*}
G^{\prime }\left( \mathbf{w}_{i}\right) =\frac{1}{\left\Vert \mathbf{w}%
_{i}\right\Vert }\mathbf{w}_{i}\quad \text{for}\quad i=1,2,...,n.
\end{equation*}%
Now from the given data, we get%
\begin{equation*}
W=\frac{2}{m\left( 1-k\right) }\left[ 
\begin{array}{cccc}
\frac{1}{\alpha _{1}} & \frac{1}{\alpha _{2}} & \cdot \cdot \cdot & \frac{1}{%
\alpha _{n}} \\ 
\frac{1}{\alpha _{1}} & \frac{1}{\alpha _{2}} & \cdot \cdot \cdot & \frac{1}{%
\alpha _{n}} \\ 
\begin{array}{c}
\cdot \\ 
\cdot%
\end{array}
& 
\begin{array}{c}
\cdot \\ 
\cdot%
\end{array}
& 
\begin{array}{c}
\cdot \\ 
\cdot%
\end{array}
& 
\begin{array}{c}
\cdot \\ 
\cdot%
\end{array}
\\ 
\frac{1}{\alpha _{1}} & \frac{1}{\alpha _{2}} & \cdot \cdot \cdot & \frac{1}{%
\alpha _{n}}%
\end{array}%
\right]
\end{equation*}%
and so we have%
\begin{equation*}
G\left( \mathbf{w}_{i}\right) =\frac{2}{\left( n-1\right) \left( 1-k\right) }%
\sqrt{\frac{1}{\alpha _{1}^{2}}+\frac{1}{\alpha _{2}^{2}}+...+\frac{1}{%
\alpha _{n}^{2}}}(1,1,...,1)\text{ for }i=1,2,...,n
\end{equation*}%
and%
\begin{equation*}
G^{\prime }\left( \mathbf{w}_{i}\right) =\frac{1}{\sqrt{\frac{1}{\alpha
_{1}^{2}}+\frac{1}{\alpha _{2}^{2}}+...+\frac{1}{\alpha _{n}^{2}}}}(\frac{1}{%
\alpha _{1}},\frac{1}{\alpha _{2}},...,\frac{1}{\alpha _{n}})\text{ for }%
i=1,2,...,n.
\end{equation*}%
It is easy to verify that $G^{\prime }$ is monotone and Lipchitz continuous
operator. The vector $\mathbf{w}=\frac{2}{m\left( 1-k\right) }(\frac{1}{%
\alpha _{1}},\frac{1}{\alpha _{2}},...,\frac{1}{\alpha _{n}})$ is the
solution of GSVM with $\left\Vert \mathbf{w}\right\Vert =\frac{2}{m\left(
1-k\right) }\sqrt{\frac{1}{\alpha _{1}^{2}}+\frac{1}{\alpha _{2}^{2}}+...+%
\frac{1}{\alpha _{n}^{2}}}.$ $\square $\newline

\noindent \textbf{Example 2.13.}\ \ \ Consider $\left( \alpha
_{1},0,0\right) ,$ $\left( 0,\alpha _{2},0\right) ,$ $\left( 0,0,\alpha
_{3}\right) ,$ $(\beta _{1},0,0),$ $(0,\beta _{2},0),$ $(0,0,\beta _{3})\ $%
as data of positive class and $\left( k\alpha _{1},0,0\right) ,$ $\left(
0,k\alpha _{2},0\right) ,$ $\left( 0,0,k\alpha _{3}\right) ,$ $(k\beta
_{1},0,0),$ $(0,k\beta _{2},0),$ $(0,0,k\beta _{3})$ as negative class of
data, where $\alpha _{i},\beta _{i}$ and $k$\ are positive real numbers with
each $\alpha _{i}\leq \beta _{i}$ for $i=1,2,3$ and $k\neq 1$.

\noindent The map $G:%
\mathbb{R}
^{n}\rightarrow 
\mathbb{R}
_{+}^{n}$ is given as%
\begin{equation*}
G\left( \mathbf{w}_{i}\right) =\left( \left\Vert \mathbf{w}_{i}\right\Vert
,\left\Vert \mathbf{w}_{i}\right\Vert ,...,\left\Vert \mathbf{w}%
_{i}\right\Vert \right) \quad \text{for}\quad i=1,2,3,
\end{equation*}%
where $\mathbf{w}_{i}$ are the row of $W_{3\times 3}$ for $i=1,2,...,n.$
Then we have%
\begin{equation*}
G^{\prime }\left( \mathbf{w}_{i}\right) =\frac{1}{\left\Vert \mathbf{w}%
_{i}\right\Vert }\mathbf{w}_{i}\quad \text{for}\quad i=1,2,3.
\end{equation*}%
Now from the given data, we get%
\begin{equation*}
W=\frac{2}{\left( 1-k\right) }\left[ 
\begin{array}{ccc}
\frac{1}{\alpha _{1}} & \frac{1}{\alpha _{2}} & \frac{1}{\alpha _{3}} \\ 
\frac{1}{\alpha _{1}} & \frac{1}{\alpha 2} & \frac{1}{\alpha _{3}} \\ 
\frac{1}{\alpha _{1}} & \frac{1}{\alpha _{2}} & \frac{1}{\alpha _{3}}%
\end{array}%
\right]
\end{equation*}%
and so we have%
\begin{equation*}
G\left( \mathbf{w}_{i}\right) =\frac{2}{\left( 1-k\right) }(\sqrt{\frac{1}{%
\alpha _{1}^{2}}+\frac{1}{\alpha _{2}^{2}}+\frac{1}{\alpha _{3}^{2}}},\sqrt{%
\frac{1}{\alpha _{1}^{2}}+\frac{1}{\alpha _{2}^{2}}+\frac{1}{\alpha _{3}^{2}}%
},\sqrt{\frac{1}{\alpha _{1}^{2}}+\frac{1}{\alpha _{2}^{2}}+\frac{1}{\alpha
_{3}^{2}}})
\end{equation*}%
and%
\begin{equation*}
G^{\prime }\left( \mathbf{w}_{i}\right) =\frac{1}{\sqrt{\frac{1}{\alpha
_{1}^{2}}+\frac{1}{\alpha _{2}^{2}}+\frac{1}{\alpha _{3}^{2}}}}(\frac{1}{%
\alpha _{1}},\frac{1}{\alpha _{2}},\frac{1}{\alpha _{3}})\text{.}
\end{equation*}%
Note that, for any $\mathbf{w}_{1},\mathbf{w}_{2}\in W,$%
\begin{equation*}
\left\Vert G^{\prime }\left( \mathbf{w}_{1}\right) -G^{\prime }\left( 
\mathbf{w}_{2}\right) \right\Vert =0=L\left\Vert \mathbf{w}_{1}-\mathbf{w}%
_{2}\right\Vert
\end{equation*}%
is satisfied for $L>0$. Also%
\begin{equation*}
<G^{\prime }\left( \mathbf{w}_{1}\right) -G^{\prime }\left( \mathbf{w}%
_{2}\right) ,\mathbf{w}_{1}-\mathbf{w}_{2}>\text{ }\geq \text{ }0
\end{equation*}%
is satisfied which show that $G^{\prime }$ is monotone operator. Moreover, $%
\mathbf{w}=\frac{2}{\left( 1-k\right) }(\frac{1}{\alpha _{1}},\frac{1}{%
\alpha _{2}},\frac{1}{\alpha 3})$ is the solution of GSVM with $\left\Vert 
\mathbf{w}\right\Vert =\frac{2}{\left( 1-k\right) }\sqrt{\frac{1}{\alpha
_{1}^{2}}+\frac{1}{\alpha _{2}^{2}}+\frac{1}{\alpha _{3}^{2}}}.$ $\square $%
\newline

\noindent \textbf{Example 2.14.}\ \ \ Let us take the group of data of
positive class $\left( 1,0,0\right) ,(1,1,0),\left( 0,1,1\right) $ and
negative class $(-\dfrac{1}{2},0,0),(-\dfrac{1}{2},-\dfrac{1}{2},0),(0,-%
\dfrac{1}{2},-\dfrac{1}{2}).$

\noindent Now from the given data, we have%
\begin{equation*}
W=\left[ 
\begin{array}{ccc}
\frac{4}{3} & 0 & \frac{4}{3} \\ 
\frac{4}{3} & 0 & \frac{4}{3} \\ 
\frac{4}{3} & 0 & \frac{4}{3}%
\end{array}%
\right]
\end{equation*}%
with%
\begin{equation*}
G\left( \mathbf{w}_{i}\right) =\frac{4}{3}(\sqrt{2},\sqrt{2},\sqrt{2})\text{
for }i=1,2,3
\end{equation*}%
and%
\begin{equation*}
G^{\prime }\left( \mathbf{w}_{i}\right) =\frac{1}{\sqrt{2}}(1,0,1)\text{ for 
}i=1,2,3.
\end{equation*}%
It is easy to verify that $G^{\prime }$ is monotone operator and Lipchitz
continuous. Moreover, $\mathbf{w}=(\dfrac{4}{3},0,\dfrac{4}{3})$ is the
solution of GSVM with $\left\Vert \mathbf{w}\right\Vert =\dfrac{4}{3}\sqrt{2}%
.$ $\square $\newline

\noindent \textbf{Conclusion}. Recently many results appeared in the
literature giving the problems related to the support vector machine and it
applications. In this paper, initiate the study of generalized support
vector machine and present linear classification of data by using support
vector machine and generalized support vector machine. We also provide
sufficient conditions under which the solution of generalized support vector
machine exist. Various examples are also present to show the validity of
these results.

\TEXTsymbol{\backslash}end\{conclusion\}

\end{document}